\documentclass[runningheads]{llncs}
\usepackage{graphicx}
\usepackage{amsmath,amssymb,amsfonts}
\usepackage{xcolor}
\usepackage{booktabs}
\usepackage{caption}
\usepackage{soul}
\usepackage{url}
\usepackage{subfig}
\begin{document}
%
%\title{AURSAD: Universal Robot Screwdriving Anomaly Detection Dataset}
%\title{Towards Automated Anomaly Detection in Screwdriving Applications: The AURSAD Dataset}
\title{Detecting Faults during Automatic Screwdriving: A Dataset and Use Case of Anomaly Detection for Automatic Screwdriving}
\titlerunning{Detecting Faults during Automatic Screwdriving: A Dataset and Use Case}
\author{Błażej Leporowski\inst{1}\and
Daniella Tola\inst{1}\and
Casper Hansen\inst{2}\and
Alexandros Iosifidis\inst{1}
}
\authorrunning{B. Leporowski et al.}
\institute{Department of Electrical and Computer Engineering, Aarhus University \and
Technicon ApS, Hobro\\
\email{\{bl,dt,ai\}@ece.au.dk}  \hspace{0.5cm}
\email{cha@techicon.dk}
}
\maketitle              % typeset the header of the contribution
\begin{abstract}
%Automated assembly is increasingly common in modern industrial applications.
%Among the various aspects of it, screwdriving stands out as one of the most common tasks. 
%As with many manufacturing operations, detecting faults can be difficult, especially if each fault model is to be engineered by hand.
Detecting faults in manufacturing applications can be difficult, especially if each fault model is to be engineered by hand.
Data-driven approaches, using Machine Learning (ML) for detecting faults have recently gained increasing interest, where a ML model can be trained on a set of data from a manufacturing process. 
In this paper, we present a use case of using ML models for detecting faults during automated screwdriving operations, and introduce a new dataset containing fully monitored and registered data from a Universal Robot and OnRobot screwdriver during both normal and anomalous operations.
%The AURSAD dataset contains 2,045 time-series with 134-dimensions.
%The structure of the dataset, its attributes, as well as the potential applications are presented in this paper.
We illustrate, with the use of two time-series ML models, how to detect faults in an automated screwdriving application.
%\hl{No idea here}
%A benchmark of fault detection, based on time-series classification, using three popular deep learning architectures is also provided and its findings are briefly discussed. 

\end{abstract}
%
%
%
%%%%%%%%%%%%%%%%%%%%%%%%%%%%%%%%%%%%%%%%%%%%%%%%%%%%%%%%%%%%%%%%%%%%%%%%%%%%%%%%%%%%%%%%%%%%%%%%%%%%%%%%%%%%%%%%%%%%%%%%%%%%%%%
\section{Introduction}
\vspace{-0.1cm}
%According to a recent survey undertaken by the ASSEMBLY magazine~\cite{assembly_survey}, screwdriving is performed at 58\% of U.S. assembly plants, which makes it more popular than welding, pressing, adhesive bonding or riveting.
%This popularity places screwdriving at the forefront of the automation drive in the industrial sector.
%Automation of such a process promises significant gains in efficiency without compromising the assembly quality.
Screwdriving is performed at 58\% of U.S. assembly plants, placing it at the forefront of the automation drive in the industrial manufacturing sector~\cite{assembly_survey}.
Automation of such a process promises significant gains in efficiency without compromising the assembly quality~\cite{automation_challenges}.
%Among factors may affect the quality of the screwing operation, such as cross-thread, sufficient torque, and correct pickup of the screws~\cite{Jia19}. 
Events such as cross-thread, sufficient torque, and correct pickup of the screws, create challenges that may affect the quality of screwdriving applications~\cite{Jia19}.
To maintain the process quality, effective means of monitoring and detecting faults need to be developed.
Creating such an automatic system or models to detect all the potential faults is challenging, due to the various failure types and different disturbances that can occur~\cite{Blanke15}.
Machine Learning (ML) has the potential to provide automated solutions for such problems, potentially eliminating or limiting the need for human experts.

By collecting more data from faulty operations a better understanding of the process critical parameters can be obtained.
The insufficient knowledge of these processes and limitations in early fault detection are a major barrier preventing deployment of more robots at lower cost in the industry.
With improved understanding and applying the ML models into robotics modules the robot operators will be able to benefit more from robotic automation in general.

The main contribution of this paper is an introduction of a novel AURSAD dataset, that we have made publicly available at~\cite{aursad}, and presentation of a fault detection use case in an automated screwdriving applications using two Deep Learning (DL) models.
By performing these experiments we illustrate the feasibility of fault detection in automated screwdriving applications.
To our knowledge this is a unique dataset that provides comprehensive information related to all available robot sensors continuously recorded throughout the whole process.

The dataset contains 2,045 samples of labeled normal and faulty data, which can be used to train DL or ML models for fault or anomaly detection. Using state-of-the-art DL models we perform fault detection, where we demonstrate the suitability of the dataset for further analysis and experimentation. 

%However, to evaluate and introduce such ML models, first an annotated dataset that describes the complete process of automated screwdriving needs to be available. 
%Current data from robots operating in the industry is mainly based on normal, fault-free operation, making it challenging to obtain and access data containing faulty operations.

%In this paper we introduce and benchmark a new dataset, called AURSAD, which is made publicly available in~\cite{aursad}.
%It contains 2,045 samples of normal and anomalous operations of the Universal Robots e-Series robot (UR3e) and OnRobot screwdriver.
%To our knowledge this is a unique dataset that provides comprehensive information related to all available robot sensors and is formed by sensor measurements continuously recorded throughout the whole process. 
%Along with the description of the dataset, we provide benchmarks showcasing the use of the dataset in multi-class classification and binary anomaly detection using state-of-the-art deep learning models recently proposed in the literature for time-series classification (TCS) problems.
%These benchmarks aim to demonstrate the suitability of the dataset for further analysis and experimentation. 
The description of the AURSAD dataset is accompanied by a technical report~\cite{aursad_arxiv}, and a Python library\footnote{\url{https://pypi.org/project/aursad/}} with functionalities that can be used for dataset preprocessing.
Additionally, code that could be used for replicating our experiments is available on GitHub\footnote{\url{https://github.com/CptPirx/AURSAD-source}}.

%The paper is organised as follows. Section~\ref{S:Relatedwork} provides comparisons to related datasets. Section~\ref{S:Dataset} describes the AURSAD dataset in detail.
%Description of the methods used to benchmark the use of the data in multi-class classification and binary anomaly detection and experimental results are provided in Section \ref{S:Experiments}.
%Finally, Section~\ref{S:Discussion} provides a summary of our work.

%%%%%%%%%%%%%%%%%%%%%%%%%%%%%%%%%%%%%%%%%%%%%%%%%%%%%%%%%%%%%%%%%%%%%%%%%%%%%%%%%%%%%%%%%%%%%%%%%%%%%%%%%%%%%%%%%%%%%%%%%%%%%%%
\vspace{-0.1cm}
\section{Background and Related Work} \label{S:Relatedwork}
\vspace{-0.1cm}
A critical analysis of public anomaly datasets presented in~\cite{Emmott13}, concluded that the most relevant attributes are: 1) point difficulty, 2) relative frequency, 3) semantic variation, and 4) feature relevance. 
% Should we discuss somewhere that we didn't check 1)? I think we checked 1 too.
The characteristics of the AURSAD dataset makes it possible to select variable amounts of anomalies, satisfying attribute 2, as well as varying cluster density, thus satisfying attribute 3. 
AURSAD allows for easy feature selection and analysis which satisfies attribute 4. Moreover, the data collection process was designed to avoid patterns that can reduce the point difficulty of the resulting time-series data, thus considering the attribute 1. 
%As can be seen, the AURSAD dataset meets the main criteria that an anomaly detection dataset should, as defined by~\cite{Emmott13}, consider. 
Section~\ref{S:Dataset} describes this in more detail.

To our knowledge, the only other available dataset for anomaly detection on screwdriving is The Manipulation Lab Screwdriving Dataset (TMLSD)~\cite{Aronson17,Cheng18}. 
Table~\ref{tab:dataset_comparison} shows a comparison of the characteristics of our AURSAD dataset and the TMLSD dataset. 
A notable difference between the two datasets is that TMLSD is more focused on each substage of the screwing process, i.e. hole finding, mating etc.\, while AURSAD focuses on the complete screwdriving process. 
Moreover, the annotations for the different categories included in the AURSAD dataset, as well as the publicly available Python library make it easy to employ in mainstream deep learning frameworks.
% \vspace{-0.1cm}
\begin{table}[htbp]
\caption{Dataset characteristics comparison}\label{tab:dataset_comparison}
\centering
    \begin{tabular}{@{}lll@{}}
    \toprule
    Dataset                                 & TMLSD~\cite{Aronson17,Cheng18}    & \textbf{AURSAD}   \\ \midrule
    Number of samples                       & 1862                              & 2045              \\
    Number of anomaly samples               & 291 (15.6\%)                      & 625 (30.5\%)      \\ 
    Number of anomalies                     & 6                                 & 4                 \\
    Number of features                      & 17                                & 125 \\
    Labels for supervised classification    & --                                & \checkmark        \\
    Publicly available                      & \checkmark                        & \checkmark        \\
    Publicly available source code          & --                                & \checkmark        \\ 
    Off-the shelf components                & --                                & \checkmark        \\  \bottomrule
    \end{tabular}
\end{table}

%%%%%%%%%%%%%%%%%%%%%%%%%%%%%%%%%%%%%%%%%%%%%%%%%%%%%%%%%%%%%%%%%%%%%%%%%%%%%%%%%%%%%%%%%%%%%%%%%%%%%%%%%%%%%%%%%%%%%%%%%%%%%%%
\vspace{-0.2cm}
\section{The AURSAD Dataset} \label{S:Dataset}
\vspace{-0.1cm}
The AURSAD dataset contains time-series sampled at 100 Hz across 2,045 samples. 
The data comes from two sources: the UR3e robot and the OnRobot screwdriver attachment. 
Technical details about the hardware and software setup can be found in our technical report~\cite{aursad_arxiv}.

The UR3e robot provides sensor data on multiple aspects of its operation.
Among them are:
\begin{itemize}
    \item target and actual joint positions, velocities, accelerations, currents and torques,
    \item target and actual Cartesian coordinates and speed of the tool, and
    \item 3-dimensional tool accelerometer values.
\end{itemize}
The OnRobot screwdriver main sensor measurements are the target and current torque and torque gradient.
In total, the screwdriver has 7 sensor features.

\vspace{-0.2cm}
\subsection{Dataset structure}\label{subsec:Dataset_structure}
\begin{figure*}[]
    \centering
    \includegraphics[width=\textwidth]{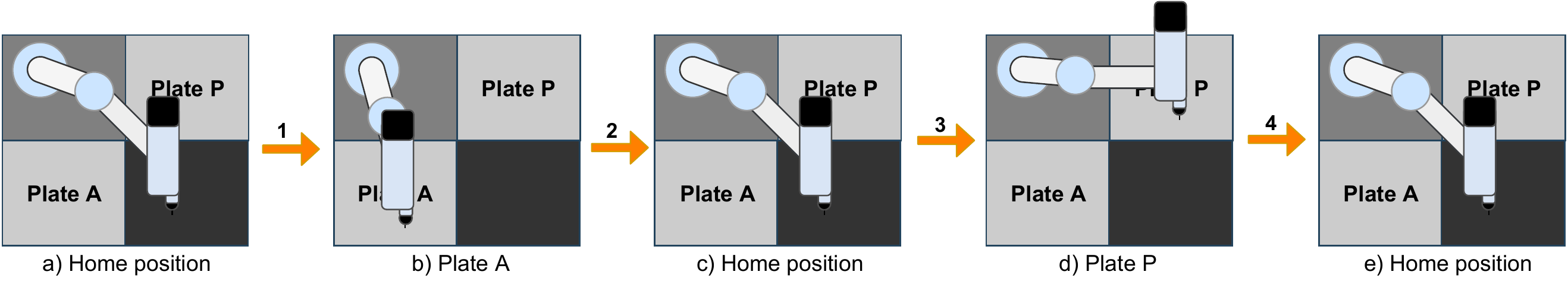}
    \caption{An example of a screw sequence: \textit{home} $\rightarrow$ \textit{plate A (loosen)}  $\rightarrow$ \textit{home} $\rightarrow$ \textit{plate P (tighten)}  $\rightarrow$ \textit{home}}
    \label{fig:sd_sketch_horizontal}
\end{figure*}

The dataset setup can be described using 3 main parts: plate A, plate P and the robot with the attached screwdriving tool.
All samples in the dataset follow the same procedure shown in Fig.~\ref{fig:sd_sketch_horizontal}.
Data was collected by recording a series of tightening and loosening movements.
On each of these recordings, one of the plates, in the example of Fig.~\ref{fig:sd_sketch_horizontal} plate A, is initially filled with screws in all its thread-holes.
The second plate, in this example plate P, initially has no screws. 
Using the auxiliary python library it is possible to label the data in different ways.
Each sample in the dataset can be chosen to consist of either the complete sequence shown in Fig.~\ref{fig:sd_sketch_horizontal}, or only subsequent parts of it. 

\vspace{-0.5cm}
\vspace{-0.5cm}

Each time-series $i$ in the dataset has a size of $t_i \times 134$ elements, where $t_i$ is the number of events of the operation sampled at~$100$ Hz and $134$ is the number of dimensions for each event. 
The duration of each time-series may vary depending on the location of the hole on the plate and the specific operation performed. 
The further away the hole used for the sample is, the longer it takes the robot to reach the destination.
Out of the 134 dimensions, 125 are measurements coming from the robot and screwdriver sensors, 8 dimensions (highlighted in bold text in Table~8 in~\cite{aursad_arxiv}) correspond to control variables hard-coded during data collection to facilitate data annotation, and one dimension indicating the label of each event.
The 8 control variable dimensions should be removed from the dataset when the data is used for ML, however, they can be useful for conducting a more in depth data analysis.

%\vspace{-0.4cm}
\subsection{Anomalies}\label{subsec:Anomalies}
%\vspace{-0.2cm}
The dataset contains 5 main types of operations plus 1 supplementary category.
The main types are:
\begin{itemize}
    \item \underline{Normal operation:} the screwdriving process is completed successfully and according to the expectations.
    \item \underline{Damaged screw anomaly:} the screw that the screwdriver picks up and then tightens has a damaged thread.
    \item \underline{Missing screw anomaly:} the screwdriver fails to pick up the screw and proceeds to the tightening stage without a screw.
    \item \underline{Extra assembly component anomaly:} during the tightening there is an additional, unexpected element (a washer) present.
    \item \underline{Damaged plate thread anomaly:} the threaded hole of the plate has been damaged. 
\end{itemize}
The damaged plate thread class is underrepresented, but can be useful for experiments with rare occurrence anomalies.
Otherwise, this class can be easily discarded from the dataset.
The supplementary type is the loosening label which describes the loosening motion.
Table \ref{tab:data_statistics} presents the distribution of the classes in the dataset.
Fig.~\ref{fig:class_means} shows the mean time-series of the \textit{current torque} sensor readings for all classes.% \hl{As can be seen, .... }

% \vspace{-0.5cm}
% \begin{figure}
%     \begin{floatrow}
%         \ffigbox{%
%           \includegraphics[width=\columnwidth]{figures/dataset_shuffle.png}%
%         }{%
%           \caption{Example of how the time-series data can be shuffled to create new datasets}%
%           \label{fig:shuffled_data}
%         }
%         \capbtabbox{
%         {\caption{Dataset statistics}%
%         \label{tab:data_statistics}
%         }
%             \begin{tabular}{@{}llll@{}}
%             \toprule
%             Type                     &  Label   &   Samples & Percentage \\ \midrule
%             Normal operation         &  0       &   1420    & 69.44 \%   \\
%             Damaged screw            &  1       &   221     & 10.81 \%    \\ 
%             Extra assembly           &  2       &   183     & 8.95 \%   \\
%             Missing screw            &  3       &   218     & 10.65 \%   \\
%             Damaged plate            &  4       &   3       & 0.15  \%   \\
%             Total                    &          &   2045    & 100 \%     \\ \bottomrule
%             \end{tabular}%
%         }{  %\caption{Dataset statistics}%
%           %\label{tab:data_statistics}
%           }
%     \end{floatrow}
% \end{figure}

\vspace{-0.5cm}
\begin{table}
	\begin{minipage}{0.47\linewidth}
		\centering
		\label{figu:shuffled_data}
		\includegraphics[width=\columnwidth]{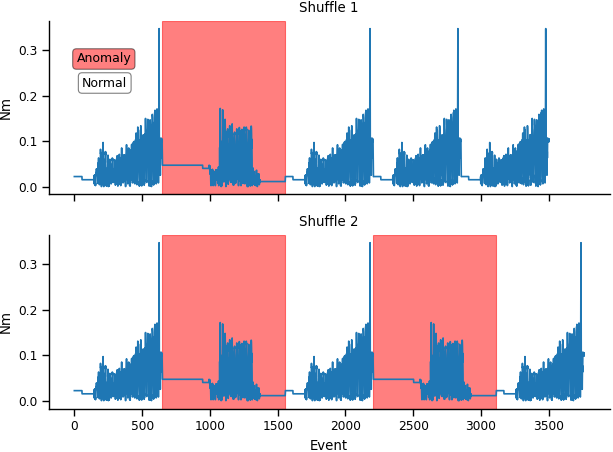}
		\captionof{figure}{Example of how the time-series data can be shuffled to create new datasets}
	\end{minipage}\hfill
	\begin{minipage}{0.47\linewidth}
		\caption{Dataset statistics}
		\label{tab:data_statistics}
		\centering
		\resizebox{\textwidth}{!}{%
		 \begin{tabular}{@{}llll@{}}
            \toprule
            Type                     &  Label   &   Samples & Percentage \\ \midrule
            Normal operation         &  0       &   1420    & 69.44 \%   \\
            Damaged screw            &  1       &   221     & 10.81 \%    \\ 
            Extra assembly           &  2       &   183     & 8.95 \%   \\
            Missing screw            &  3       &   218     & 10.65 \%   \\
            Damaged plate            &  4       &   3       & 0.15  \%   \\
            Total                    &          &   2045    & 100 \%     \\ 
            \bottomrule
            \end{tabular}}
	\end{minipage}\hfill
\end{table}

% \begin{table}[h]
% \caption{Dataset statistics}
% \centering
%     \begin{tabular}{@{}llll@{}}
%     \toprule
%     Type                     &  Label   &   Samples & Percentage \\ \midrule
%     Normal operation         &  0       &   1420    & 69.44 \%   \\
%     Damaged screw            &  1       &   221     & 10.81 \%    \\ 
%     Extra assembly component &  2       &   183     & 8.95 \%   \\
%     Missing screw            &  3       &   218     & 10.65 \%   \\
%     Damaged plate thread     &  4       &   3       & 0.15  \%   \\
%     Total                    &          &   2045    & 100 \%     \\ \bottomrule
%     \end{tabular}
% \label{tab:data_statistics}

% \end{table}
% \begin{figure}
%     \centering
%     \includegraphics[width=0.5\columnwidth]{figures/Class_means_linear_sns.png}
%     \caption{Mean time-series of current torque measurements for all classes in the AURSAD dataset.}
%     \label{fig:class_means}
% \end{figure}

% \vspace{-0.5cm}
\subsection{Dataset applicability}\label{subsec:Dataset_usage_areas}
The main target of the dataset is anomaly and fault detection using ML.
The sequence of each operation was chosen specifically to start and end in the home position, as indicated in Fig.~\ref{fig:sd_sketch_horizontal}, to allow users of the dataset to mix the data in a different order and experiment with different numbers of anomalies when training, validating and testing. 
This gives the users flexibility, while at the same time fulfilling the attributes defined by~\cite{Emmott13} which state that an anomaly detection dataset must have a relative frequency, and semantic variation. 
Fig.~\ref{fig:shuffled_data} gives examples of how the time-series data can be shuffled into different datasets, depending on the user's needs.

The dataset also contains boolean register flags, indicating when the robot moves to home position, performs a tightening operation or a loosening operation. 
These flags provide the users with data to determine the trajectory of the robot, its joint angles etc., meaning that this data can also be used during the creation of models of the UR3e robot or the Onrobot screwdriver.

%%%%%%%%%%%%%%%%%%%%%%%%%%%%%%%%%%%%%%%%%%%%%%%%%%%%%%%%%%%%%%%%%%%%%%%%%%%%%%%%%%%%%%%%%%%%%%%%%%%%%%%%%%%%%%%%%%%%%%%%%%%%%%%
\vspace{-0.1cm}
\section{Experiments} \label{S:Experiments}
\vspace{-0.1cm}
In this Section, we provide a benchmark of the use of AURSAD dataset for multi-class classification and binary anomaly detection problems. 
The experiments performed in this paper are based solely on the tightening and movement operation, consisting of sub-sequence \textit{c, d} and~\textit{e} illustrated in Fig.~\ref{fig:sd_sketch_horizontal}. 
The loosening and damaged plate thread classes have been excluded from the experiments because of their irrelevance and small sample count, respectively. 
We use all 125 features of the dataset. 
To effectively perform the fault classification experiments with different deep learning model architectures, the time-series data have been zero padded to achieve identical lengths. 
We used the following deep learning models in our experiment:

\begin{figure}[!htbp]
    \centering
    \includegraphics[width=0.7\columnwidth]{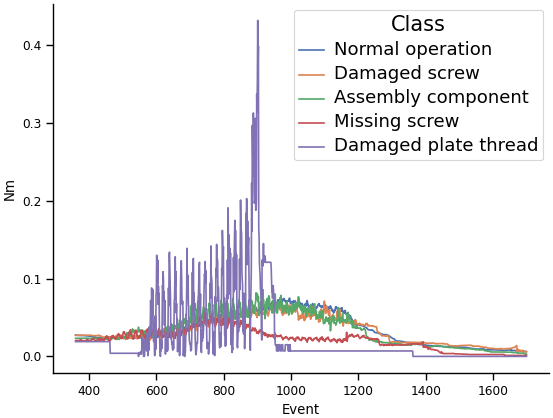}
    \caption{Mean time-series of current torque measurements for all classes in the AURSAD dataset}
    \label{fig:class_means}
\end{figure}

\paragraph{Residual Neural Network:} 
    According to the review of deep learning methods for TSC in~\cite{tsc_review}, ResNet~\cite{resnet} architecture is consistently among the highest scoring ones across different datasets, and is therefore used in our experiments.
    The implementation of the network we used in our experiments is based on the publicly available repository of the TSC review. 
    It consists of 3 blocks employing 1-dimensional convolution with Rectified Linear Unit activation and batch normalization. 
    
\paragraph{Temporal Attention-augmented Bilinear Network (TABL)}: 
    The TABL~\cite{tabl} network is based on bilinear layers (BL) with introduced attention mechanism along the time dimension.
    The method was designed for use with time-series, and the BLs are able to learn two separate dependencies for the two modes of a multivariate data.
    % One transformation is learned for the input dimensions and the second aggregates information in the time domain.
    % The addition of an attention mechanism to the last layer allows the network to focus on the most impactful time instances of the input time-series, encouraging competition between neurons that represent different temporal steps of the same feature.
    We have performed a small hyperparameter search, and the variant of TABL network used in our experiments contains 3 BL with shapes [240, 5], [120, 2] and [60, 1] and a TABL layer of shape [4, 1].

In all our experiments, we used used a split of 70/30\% of data for training and testing the neural networks, respectively. Training was conducted for $100$ epochs with learning rate reduction on plateau, drop rate varying from 10\% to 30\% and using the Adam optimizer.
We used the averaged F1 score to measure the performance of each method, as well as measuring per class F1 score. 
For each experiment the networks have been trained and evaluated 3 times and the averaged performance is reported. 

The results for the TABL and ResNet classifiers are shown in Table~\ref{tab:tabl_results} and Table~\ref{tab:resnet_results}, respectively.

\vspace{-0.5cm}
\begin{table}
%\RawFloats
    \parbox{.45\linewidth}{
        \centering
        \caption{TABL performance}
        \label{tab:tabl_results}
        \begin{tabular}{@{}llll@{}}
            \toprule
            Label   & Precision         & Recall            & F1        \\ \midrule
            0       & 0.839             & 0.488             & 0.617     \\
            1       & 0.183             & 0.455             & 0.261     \\
            2       & 0.179             & 0.436             & 0.254     \\
            3       & 0.881             & 0.894             & 0.887     \\ \midrule
            Average & 0.520             & 0.568             & 0.505     \\  \bottomrule
        \end{tabular}
    }
    \hfill
    \parbox{.45\linewidth}{
    \caption{ResNet performance}
            \label{tab:resnet_results}
        \centering
            \begin{tabular}{@{}llll@{}}
                \toprule
                Label   & Precision     & Recall    & F1         \\ \midrule
                0       & 0.913         & 0.887     & 0.9        \\
                1       & 0.48          & 0.546     & 0.511      \\
                2       & 0.786         & 0.8       & 0.793      \\
                3       & 0.927         & 0.955     & 0.94       \\ \midrule
                Average & 0.776         & 0.797     & 0.789      \\ \bottomrule
            \end{tabular}
            
    }
\end{table}
\vspace{-0.5cm}

Fig.~\ref{fig:class_means} shows the missing screw anomaly is distinct compared to other types of operations.
It could therefore be expected that the models distinguish it from the other classes with good confidence.
The experiment results reaffirm this observation, as both tested models achieved good performance on this anomaly.
The normal operation class was also classified well by both of the tested models, which can probably be attributed to its large representation in the dataset. 
Differences between the extra assembly component class and the damaged screw class are more subtle, and the smaller, less complicated TABL model struggled to recognise those classes.

To check if the robot features provide meaningful data to distinguish the classes, a comparison between the full dataset and a limited amount of dimensions containing only the 7 features directly sampled from the screwdriver itself has been performed.
The performance achieved by each deep learning model is shown in Table~\ref{tab:screwdriver_only_results}. 
Comparing the results in Table~\ref{tab:screwdriver_only_results} with those in Tables~\ref{tab:resnet_results} and \ref{tab:tabl_results} it can be seen that the models performed worse when using the sensor measurements directly sampled from the screwdriver itself. 
The TABL performs better than ResNet, which can probably be attributed to the more complicated ResNet model overfitting the now reduced dataset. 
The smaller amount of features alleviates the underfitting problem of the TABL model, hence it performs better on the reduced dataset.

\vspace{-0.5cm}
\begin{table}
%    \RawFloats
    \parbox{.45\linewidth}{
    \centering
        \caption{Screwdriver subset F1 results}
        \begin{tabular}{@{}lll@{}}
            \toprule
            \text{Label} & \multicolumn{2}{c@{}}{\text{Classifier}}\\
            \cmidrule(l){2-3}
                    & ResNet        & TABL        \\
            0       & 0.773         & 0.85        \\
            1       & 0.34          & 0.413         \\
            2       & 0.424         & 0.417          \\
            3       & 0.927         & 0.931        \\ \midrule
            Average & 0.616         & 0.653        \\ \bottomrule
        \end{tabular}
        \label{tab:screwdriver_only_results}
    }
    \hfill
    \parbox{.45\linewidth}{
    \centering
        \caption{Binary classification F1 results}
        \begin{tabular}{@{}lll@{}}
            \toprule
            \text{Label} & \multicolumn{2}{c@{}}{\text{Classifier}}\\
            \cmidrule(l){2-3}
                    & ResNet        & TABL        \\
            Normal  & 0.91          & 0.86         \\
            Anomaly & 0.8           & 0.532         \\ \midrule
            Average & 0.855         & 0.696          \\ \bottomrule
        \end{tabular}
        \label{tab:binary_results}
    }
\end{table}

% \begin{table}[h]
% \centering
% \caption{Screwdriver subset F1 results}
% \begin{tabular}{@{}llll@{}}
%     \toprule
%     \text{Label} & \multicolumn{3}{c@{}}{\text{Classifier}}\\
%     \cmidrule(l){2-4}
%             & ResNet        & TABL       & CLSTM        \\
%     0       & 0.773         & 0.85       & 0.786         \\
%     1       & 0.34          & 0.413      & 0.368         \\
%     2       & 0.424         & 0.417      & 0.31          \\
%     3       & 0.927         & 0.931      & 0.916         \\ \midrule
%     Average & 0.616         & 0.653      & 0.595         \\ \bottomrule
% \end{tabular}
% \label{tab:screwdriver_only_results}
% \end{table}
\vspace{-0.5cm}
Finally, we tested the case of binary anomaly detection, where all classes corresponding to an anomaly are merged to form one anomaly class that needs to be distinguished from the normal class. 
The performance of the models is provided in Table~\ref{tab:binary_results}. 
The ResNet network has achieved a good averaged F1 score of 0.855 on the binary classification problem.
The 0.8 F1 score for the anomaly conglomerate class shows that the anomalies in the AURSAD, without distinction between their types, can be distinguished from normal operation with high accuracy.
The difference most likely stems from the binary classes being more balanced in terms of number of samples.

% \begin{table}[h]
% \centering
% \caption{Binary classification F1 results}
% \begin{tabular}{@{}llll@{}}
%     \toprule
%     \text{Label} & \multicolumn{3}{c@{}}{\text{Classifier}}\\
%     \cmidrule(l){2-4}
%             & ResNet        & TABL       & CLSTM        \\
%     Normal  & 0.91          & 0.86       & 0.868         \\
%     Anomaly & 0.8           & 0.532      & 0.697         \\ \midrule
%     Average & 0.855         & 0.696      & 0.782          \\ \bottomrule
% \end{tabular}
% \label{tab:binary_results}
% \end{table}

%%%%%%%%%%%%%%%%%%%%%%%%%%%%%%%%%%%%%%%%%%%%%%%%%%%%%%%%%%%%%%%%%%%%%%%%%%%%%%%%%%%%%%%%%%%%%%%%%%%%%%%%%%%%%%%%%%%%%%%%%%%%%%%
\vspace{-0.1cm}
\section{Concluding Remarks}  \label{S:Discussion}

This paper introduced a time-series dataset focused on anomaly detection in automated screwdriving based on machine learning.
The AURSAD dataset has been created with the critique of previous time-series datasets~\cite{Emmott13} in mind and meets the criteria set out to rectify issues commonly appearing in time-series datasets.
The dataset contains the full range of robot and screwdriver sensor data for the whole operation of picking up/loosening the screws, movement to position and tightening, which potentially makes it also useful for tasks included in the modelling of such procedures.
We also provided benchmarks based on well established deep learning models for time-series classification and showed that the AURSAD dataset has good potential to facilitate research for fault and anomaly detection in time-series. 

The current limitations of the AURSAD dataset and our use case approach are the limited amount of samples in the damaged plate thread class and arbitrary choice of the supposedly most important features. 
In the future it may be worthwhile to consider other approaches for determining the most relevant features, such as explainable AI methods, which could help to identify the most important features as well as highlight potential issues.

\vspace{-0.1cm}
\section*{Acknowledgment}  \label{S:Acknowledgment}
This work is supported by the Smart Industry project (Grant No. RFM-17-0020) granted by the EU Regional Development Fund.

\bibliographystyle{splncs04}
\bibliography{refs}

\end{document}